\documentclass[11pt,a4paper]{article}
\usepackage[hyperref]{acl2018}
\usepackage{times}
\usepackage{latexsym}
\usepackage{url}
\usepackage{boldline}
\usepackage{array}
\usepackage{multirow}
\usepackage{graphicx}
\newcolumntype{C}[1]{>{\centering\arraybackslash}m{#1}}

\aclfinalcopy % Uncomment this line for the final submission
\def\aclpaperid{7} %  Enter the acl Paper ID here

\title{Extrofitting: Enriching Word Representation and its Vector Space \\
with Semantic Lexicons}

\author{Hwiyeol Jo \\
  AI Lab, LG Electronics \\
  {\tt hwiyeolj@gmail.com} \\\And
    Stanley Jungkyu Choi\\
  AI Lab, LG Electronics \\
  {\tt stanley.choi@lge.com} \\
}
\date{2018.04.11}

\begin{document}
\maketitle
\begin{abstract}

    We propose post-processing method for enriching not only word representation but also its vector space using semantic lexicons, which we call {\em extrofitting}. The method consists of 3 steps as follows: (i) Expanding 1 or more dimension(s) on all the word vectors, filling with their representative value. (ii) Transferring semantic knowledge by averaging each representative values of synonyms and filling them in the expanded dimension(s). These two steps make representations of the synonyms close together. (iii) Projecting the vector space using Linear Discriminant Analysis, which eliminates the expanded dimension(s) with semantic knowledge. When experimenting with GloVe, we find that our method outperforms Faruqui's retrofitting on some of word similarity task. We also report further analysis on our method in respect to word vector dimensions, vocabulary size as well as other well-known pretrained word vectors (e.g., Word2Vec, Fasttext).
    
\end{abstract}

\section{Introduction}

    As a method to represent natural language on computer, researchers have utilized distributed word representation. The distributed word representation is to represent a word as n-dimensional float vector, hypothesizing that some or all of the dimensions may capture semantic meaning of the word. The representation has worked well in various NLP tasks, substituting one-hot representation~\cite{turian2010word}. Two major algorithms learning the distributed word representation are CBOW (Continuous Bag-of-Words) and skip-gram~\cite{mikolov2013distributed}. Both CBOW and skip-gram learn the representation using one hidden neural networks. The difference is that CBOW learns the representation of a center word from neighbor words whereas skip-gram gets the representation of neighbor words from a center word. Therefore, the algorithms have to depend on word order, because their objective function is to maximize the probability of occurrence of neighbor words given the center word. Then a problem occurs because the word representations do not have any information to distinguish synonyms and antonyms. For example, {\tt worthy} and {\tt desirable} should be mapped closely on the vector space as well as {\tt agree} and {\tt disagree} should be mapped apart, although they occur on a very similar pattern. Researchers have focused on the problem, and their main approaches are to use semantic lexicons~\cite{faruqui2014retrofitting, mrkvsic2016counter, speer2017conceptnet, vulic2017cross, camacho2015nasari}. One of the successful works is Faruqui's retrofitting\footnote{The retrofitting codes are available at\\ \url{https://github.com/mfaruqui/retrofitting}}, which can be summarized as pulling word vectors of synonyms close together by weighted averaging the word vectors on a fixed vector space (it will be explained in Section~\ref{sec:2.1}). The retrofitting greatly improves word similarity between synonyms, and the result not only corresponds with human intuition on words but also performs better on document classification tasks with comparison to original word embeddings~\cite{kiela2015specializing}. From the idea of retrofitting, our method hypothesize that we can enrich not only word representation but also its vector space using semantic lexicons\footnote{Our codes are available at\\ \url{https://github.com/HwiyeolJo/Extrofitting}}. We call our method as {\em extrofitting}, which retrofits word vectors by expanding its dimensions.
    %We first describe the background of retrofitting and Linear Discriminant Analysis, which is used for dimension reduction in our method. In Section~\ref{sec:3}, we will introduce experiment data including pretrained word vectors, semantic lexicons, and word similarity dataset. After that, we will describe our method and experimental results on word similarity tasks.
    %\url{http://github.com/HwiyeolJo/Extrofitting}.

\section{Backgrounds}
\subsection{Retrofitting}
    \label{sec:2.1}
    Retrofitting \cite{faruqui2014retrofitting} is a post-processing method to enrich word vectors using synonyms in semantic lexicons. The algorithm learns the word embedding matrix $Q = \{q_1 , q_2 , \dots , q_n\}$ with the objective function $\Psi(Q)$:
    \begin{equation}
    \Psi (Q) = \sum_{i=1}^n \ [\alpha || q_i - \hat{q}_i ||^2 + \sum_{(i,j) \in E} \beta_{ij} ||q_i - q_j||^2] 
    \end{equation}
    where an original word vector is ${q}_i$, its synonym vector is $q_j$, and inferred word vector is $\hat{q}_i$. The hyperparameter $\alpha$ and $\beta$ control the relative strengths of associations. The $\hat{q}_i$ can be derived by the following online update:
    $\hat{q}_i = \frac{\sum_{j:(i,j) \in E} \beta_{ij} q_j + \alpha_i q_i}{\sum_{j:(i,j) \in E} \beta_{ij} + \alpha_i}$

\subsection{Linear Discriminant Analysis (LDA)}
    
    LDA~\cite{welling2005fisher} is one of the dimension reduction algorithms that project data into different vector space, while minimizing the loss of class information as much as possible. As a result, the algorithm finds linear vector spaces which minimize the distance of data in the same class as well as maximize the distance among the different class. The algorithm can be summarized as follows:\\
    \textbf{Calculating between-class scatter matrix $S_B$ and within-class scatter matrix $S_W$.} \\
    When we denote data as $x$, classes as $c$, $S_B$ and $S_W$ can be formulated as follows:
    \begin{equation}
    S_B = \sum_{c} (\mu_i - \mu) (\mu_i - \mu)^T,
    \end{equation}
    \begin{equation}
    S_{W} = \sum_{c} \sum_{i \in c} (x_i - \mu_c) (x_i - \mu_c)^T,
    \end{equation}
    where the overall average of $x$ is $\mu$, and the partial average in class $i$ is denoted by $\mu_i$.\\
    \textbf{Maximizing the objective function $J(w)$.}\\
    The objective function $J(w)$ that we should maximize can be defined as
    \begin{equation}
    J(w) = \frac{| U^T S_B U |}{| U^T S_W U |},    
    \end{equation}
    and its solution can be reduced to find U that satisfies $S_W^{-1} S_B = U \Lambda U^T.$
    Therefore, $U$ is derived by eigen-decomposition of $S_{W_i}^{-1} S_B$; choosing $q$ eigen vectors which have the top-$q$ eigen values, and composing transform matrix of $U$.\\
    \textbf{Transforming data onto new vector space}\\
    Using transform matrix $U$, we can get transformed data by $y = U^T x$

    \begin{table*}[h] \centering
    \begin{tabular}{|l||C{1.5cm}|C{1.5cm}|C{1.5cm}|C{1.5cm}||C{1.8cm}|C{1.2cm}|}
    \hlineB{3}
        & \bf MEN-3k & \bf WS353 & \bf SL-999 & \bf RG-65 & \#Extrofitted & \#Vocab. \\
    \hlineB{3}
    {\small\tt glove.6B.300d}    & 0.7486 & 0.5170 & 0.3705 & 0.7693 & - & 0.4M \\
    \hline
    + PPDB  & \bf 0.7949 & 0.5826 & 0.4387 & 0.8177 & 67,729 & - \\
    \hline
    + WordNet$_{syn}$  & 0.7884 & 0.5805 & \bf 0.4409 & 0.7943 & 55,388 & - \\ 
    \hline
    + WordNet$_{all}$   & 0.7893 & 0.5714 & 0.4353 & 0.8010 & 55,388 & - \\
    \hline
    + FrameNet  & 0.7840 & \bf 0.5837 & 0.4376 & \bf 0.8187 & 7,592 & - \\
    \hlineB{2}
    {\small\tt glove.42B.300d}    & 0.7435 & 0.5516 & 0.3738 & 0.8172 & - & 1.9M \\
    \hline
    + PPDB  & \bf 0.8292 & 0.6613 & \bf 0.4896 & 0.8362 & 76,631 & - \\
    \hline
    + WordNet$_{syn}$  & 0.8230 & 0.6605 & 0.4884 & \bf 0.8634 & 70,411 & - \\ 
    \hline
    + WordNet$_{all}$   & 0.8223 & \bf 0.6638 & 0.4858 & 0.8561 & 70,411 & - \\
    \hline
    + FrameNet  & 0.8123 & 0.6448 & 0.4601 & 0.8556 & 7,809 & - \\
    \hlineB{3}
    \end{tabular}
    \caption{Spearman's correlation of extrofitted word vectors for word similarity tasks using semantic lexicon. Our method improves pretrained GloVe in different vocabulary size.}
    \label{tab:1}
    \end{table*}
    
\section{Enriching Representations of Word Vector and The Vector Space}

\subsection{Expanding Word Vector with Enrichment}
    We simply enrich the word vectors by expanding dimension(s) that add 1 or more dimension to original vectors, filling with its representative value $r_i$, which can be a mean value. We denote an original word vectors as $q_i = (e_1 , e_2 , \cdots , e_D)$ where D denotes the number of word vector dimension. Then, the representative value $r_i$ can be formulated as $r_i = mean (e_1 , e_2 , \cdots , e_D)$. Intuitively, if we expand more additional dimensions, the word vectors will strengthen its own meaning. Likewise, the ratio of the number of expanded dimension to the number of original dimensions will affect the meaning of the word vectors.
\subsection{Transferring Semantic Knowledge}
    To transfer semantic knowledge on the representative value $r_i$, we also take a simple approach of averaging all the representative values of each synonym pair, substituting each of its previous value. We get the synonym pairs from lexicons we introduced in Section 3. The transferred representative value $\bar{r}_i$ can be formulated as $\bar{r}_i = \sum_{s \in L} r_s / N$ where L refers to the lexicon consisting of synonym pairs $s$, and $N$ is the number of synonyms. This manipulation makes the representation of the synonym pairs close to one another.
\subsection{Enriching Vector Space}
    With the enriched vectors and the semantic knowledge, we perform Linear Discriminant Analysis for dimension reduction as well as clustering the synonyms from semantic knowledge. LDA finds new vector spaces to cluster and differentiate the labeled data, which are synonym pairs in this experiment. We can get the extrofitted word embedding matrix $\bar{w}$ as follows:
    \begin{equation}
    \bar{Q} = \textit{LDA}( Q \oplus \bar{r}, l )
    \end{equation}
    where $Q$ is the word embedding matrix composed of word vectors $q$ and $l$ is the index of the synonym pair. We implement our method using Python2.7 with scikit-learn~\cite{scikit-learn}.

\section{Experiment Data}
\subsection{Pretrained Word Vectors}
    \label{pretrainedwordvectors}
    \textbf{GloVe}~\cite{pennington2014glove} has lots of variations in respect to word dimension, number of tokens, and train sources. We used {\small\tt glove.6B} trained on Wikipedia+Gigawords and {\small\tt glove.42B.300d} trained on Common Crawl. The other pretrained GloVe do not fit in our experiment because they have different word dimension or are case-sensitive.
    %, which is developed from Google, also train the word vectors using the co-occurrence.
    % The difference is that the dot products of word vectors equal the cosine similarity between two words whereas GloVe equals the logarithm of the words' probability of co-occurrence.
    We also use 300-dimensional \textbf{Word2Vec}~\cite{mikolov2013efficient} with negative sampling trained on GoogleNews corpus.
    \textbf{Fasttext}~\cite{bojanowski2016enriching} is an extension of Word2Vec, which utilizes subword information to represent an original word. We used 300-dimensional pretrained Fasttext trained on Wikipedia ({\small\tt wiki.en.vec}), using skip-gram.
    
\subsection{Semantic Lexicons}
    We borrow the semantic lexicons from retrofitting~\cite{faruqui2014retrofitting}.
    \citeauthor{faruqui2014retrofitting} extracted the synonyms from \textbf{PPDB}~\cite{ganitkevitch2013ppdb} by finding a word that more than two words in another language are corresponding with. Retrofitting also used \textbf{WordNet}~\cite{miller1995wordnet} database which grouped words into set of synonyms (synsets). We used two versions of WordNet lexicon, one which consists of synonym only (WordNet$_{syn}$) and the other with additional hypernyms, hyponyms included (WordNet$_{all}$). Lastly, synonyms were extracted from \textbf{FrameNet}~\cite{baker1998berkeley}, which contains more than 200,000 manually annotated sentences linked to semantic frames. \citeauthor{faruqui2014retrofitting} regarded words as synonyms if the words can be grouped with any of the frames.

\subsection{Evaluation Data}
    We evaluate our methods on word similarity tasks using 4 different kinds of dataset. {\bf MEN-3k}~\cite{bruni2014multimodal} consists of 3000-word pairs rated from 0 to 50. {\bf WordSim-353}~\cite{finkelstein2001placing} consists of 353-word pairs rated from 0 to 10. {\bf SimLex-999}~\cite{hill2015simlex} includes 999-word pairs rated from 0 to 10. {\bf RG-65}~\cite{rubenstein1965contextual} has 65 words paired scored from 0 to 4. MEN-3k and WordSim-353 were split into train (or dev) set and test set, but we combined them together solely for evaluation purpose.
    % \textbf{MEN-3k} The MEN dataset consists of 3,000 word pairs, randomly selected from words that occur at least 700 times in the freely available ukWaC and Wackypedia corpora combined (size: 1.9B and 820M tokens, respectively) and at least 50 times (as tags) in the open source subset of the ESP game dataset.\\
    % \textbf{WS353} contains a list of all 353 words, along with their mean similarity scores.\\
    % \textbf{SimLex-999} SimLex-999 comprises 666 Noun-Noun pairs, 222 Verb-Verb pairs and 111 Adjective-Adjective pairs.  measuring how well models capture similarity, rather than relatedness or association. \\
    % \textbf{RG-65} 65 word pairs. \\
    The other datasets have lots of out-of-vocabulary, so we disregard them for future work.

    \begin{table*}[h] \centering
    \begin{tabular}{|c||C{1.5cm}|C{1.5cm}|C{1.5cm}|C{1.5cm}||C{1.6cm}|}
    \hlineB{3}
        & \bf MEN-3k & \bf WS353 & \bf SL-999 & \bf RG-65 & Lexicon \\
    \hlineB{3}
    {\small\tt glove.6B.50d} & 0.6574 & 0.4193 & 0.2646 & 0.5948 & - \\
    \hline
    + Retrofitting  & 0.6773 & 0.4121 & \bf 0.3761 & \bf 0.7027 & WordNet$_{all}$ \\
    \hline
    + Extrofitting  & \bf 0.6876 & \bf 0.4859 & 0.2926 & 0.6743 & WordNet$_{all}$ \\ 
    \hlineB{2}
    {\small\tt glove.6B.100d} & 0.6932 & 0.4488 & 0.2975 & 0.6762 & - \\
    \hline
    + Retrofitting  & 0.7052 & 0.4428 & \bf 0.4065 & \bf 0.7863 & WordNet$_{all}$ \\
    \hline
    + Extrofitting  & \bf 0.7447 & \bf 0.5337 & 0.3733 & 0.7341 & WordNet$_{all}$ \\ 
    \hlineB{2}
    {\small\tt glove.6B.200d}   & 0.7244 & 0.4866 & 0.3403 & 0.7128 & - \\
    \hline
    + Retrofitting  & 0.7397 & 0.4799 & \bf 0.4415 & \bf 0.8123 & WordNet$_{all}$ \\
    \hline
    + Extrofitting  & \bf 0.7689 & \bf 0.5416 & 0.4120 & 0.7389 & WordNet$_{all}$ \\ 
    \hlineB{2}
    {\small\tt glove.6B.300d}   & 0.7486 & 0.5130 & 0.3705 & 0.7693 & - \\
    \hline
    + Retrofitting  & 0.7681 & 0.5232 & \bf 0.4701 & \bf 0.8499 & WordNet$_{all}$ \\
    \hline
    + Extrofitting  & \bf 0.7893 & \bf 0.5714 & 0.4353 & 0.8010 & WordNet$_{all}$ \\ 
    \hlineB{3}
    \end{tabular}
    \caption{Comparison of Spearman's correlation of retrofitted or extrofitted word vectors for word similarity tasks. Our method, extrofitting, outperforms retrofitting on MEN-3k and WordSim-353.}
    \label{tab:2}
    \end{table*}
    
    \begin{figure*}[!]
    \includegraphics[scale=0.5]{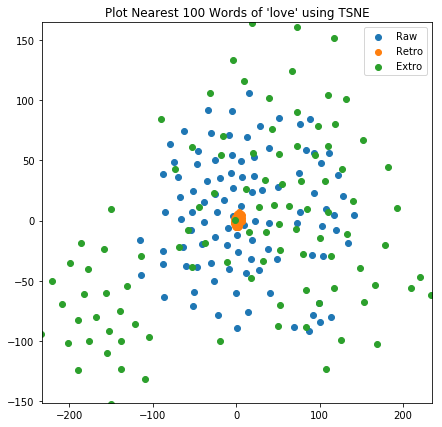}
    \includegraphics[scale=0.5]{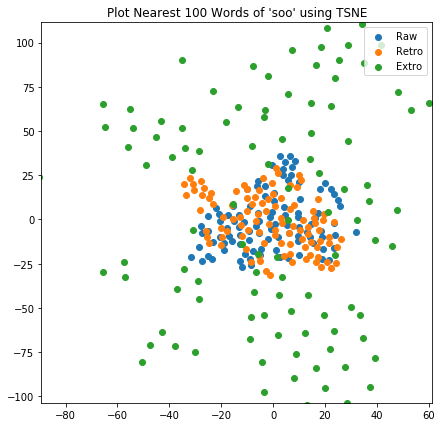}
    \caption{Plots of nearest top-100 words of cue words in different post-processing methods. We choose two cue words; one is included in semantic lexicons ({\tt love}; left), and another is not ({\tt soo}; right)}
    \label{fig:1}
    \end{figure*}
    
    \begin{table*}[h] \centering
    \begin{tabular}{|c||c|c|}
    \hlineB{3}
        Cue Word & Method & Top-10 Nearest Words(Cosine Similarity Score) \\
    \hlineB{3}
    \multirow{5}{*}{{\tt love}}   & {\small\tt glove.42B.300d} &
        \begin{tabular}{@{}c@{}c@{}}
             {\small loved(.7745), i(.7338), loves(.7311), know(.7286), loving(.7263),}\\
             {\small really(.7196), always(.7193), want(.7192), hope(.7127), think(.7110)}
        \end{tabular} \\
        \cline{2-3}
                            & {+ Retrofitting} &
        \begin{tabular}{@{}c@{}c@{}}
             {\small loved(.7857), know(.7826), like(.7781), want(.7736), i(.7707),}\\
             {\small feel(.7549), wish(.7549), think(.7491), enjoy(.7453), loving(.7451)}
        \end{tabular} \\
        \cline{2-3}
                            & {+ Extrofitting} &
        \begin{tabular}{@{}c@{}c@{}}
             {\small loved(.6008), adore(.5949), hate(.5949), luv(.5562), loving(.5391),}\\
             {\small loooove(.5321), looooove(.5233), loveeee(.5195), want(.5171), looove(.5107)}
        \end{tabular} \\
        \cline{2-3}
    \hline
    \multirow{6}{*}{{\tt soo}}    & {\small\tt glove.42B.300d} &
        \begin{tabular}{@{}c@{}c@{}}
             {\small sooo(.8394), soooo(.7938), sooooo(.7715), soooooo(.7359), sooooooo(.6844),}\\
             {\small haha(.6574), hahah(.6320), damn(.6247), omg(.6244), hahaha(.6219)}
        \end{tabular} \\
        \cline{2-3}
                            & {+ Retrofitting} &
        \begin{tabular}{@{}c@{}c@{}}
             {\small sooo(.8394), soooo(.7938), sooooo(.7715), soooooo(.7359),}\\
             {\small haha(.6574), hahah(.6320), omg(.6244), hahaha(.6219), sooooooo(.6189)}
        \end{tabular} \\
        \cline{2-3}
                            & {+ Extrofitting} &
        \begin{tabular}{@{}c@{}c@{}}
             {\small sooo(.8329), soooo(.7896), sooooo(.7774), soooooo(.7560), sooooooo(.7256),}\\
             {\small soooooooo(.6867), sooooooooo(.6796),
             soooooooooo(.6517),}\\
             {\small tooo(.6493), sooooooooooo(.6423)}
        \end{tabular} \\
        \cline{2-3}
    \hlineB{3}
    \end{tabular}
    \caption{List of top-10 nearest words of cue words in different post-processing methods. We show cosine similarity scores of two words included in semantic lexicon ({\tt love}) or not ({\tt soo}).}
    \label{tab:3}
    \end{table*}
    
    \begin{table*}[h] \centering
    \begin{tabular}{|l||C{1.5cm}|C{1.5cm}|C{1.5cm}|C{1.5cm}||C{1.8cm}|C{1.2cm}|}
    \hlineB{3}
        & \bf MEN-3k & \bf WS353 & \bf SL-999 & \bf RG-65 & \#Extrofitted & \#Vocab. \\
    \hlineB{3}
    % glove.42B.300d    & 0.7435 & 0.5516 & 0.3738 & 0.8172 & - & 1.9M \\
    % \hline
    % + PPDB  & 0.7483 & 0.5408 & 0.3943 & 0.8173 & 76,631 & - \\
    % \hline
    % + WordNet$_{syn}$  & 0.7762 & 0.5834 & 0.4165 & 0.8192 & 70,411 & - \\ 
    % \hline
    % + WordNet$_{all}$   & 0.7783 & 0.5834 & 0.4180 & 0.8188 & 70,411 & - \\
    % \hline
    % + FrameNet  & \bf 0.7872 & \bf 0.5986 & \bf 0.4269 & \bf 0.8207 & 7,809 & - \\
    % \hlineB{0.8pt}
    {\small\tt w2v-google-news} & 0.7764 & \bf 0.6156 & 0.4475 & 0.7558 & - & 3.0M \\
    \hline
    + PPDB  & \bf 0.7883 & 0.5935 & \bf 0.4799 & \bf 0.7877 & 63,825 & - \\
    \hline
    + WordNet$_{syn}$  & 0.7821 & 0.6004 & 0.4741 & 0.7844 & 64,248 & - \\ 
    \hline
    + WordNet$_{all}$   & 0.7782 & 0.6051 & 0.4733 &  0.7782 & 64,248 & - \\
    \hline
    + FrameNet  & 0.7784 & 0.6025 & 0.4651 & 0.7650 & 7,559 & - \\
    \hlineB{2}
    {\small\tt wiki.en.vec} & 0.7654 & 0.6301 & 0.3803 & \bf 0.8005 & - & 2.5M \\
    \hline
    + PPDB  & \bf 0.7737 & 0.6363 & 0.4133 & 0.7723 & 69,237 & - \\
    \hline
    + WordNet$_{syn}$  & 0.7599 & 0.6326 & \bf 0.4135 & 0.7633 & 70,542 & - \\ 
    \hline
    + WordNet$_{all}$   & 0.7569 & \bf 0.6421 & 0.4093 & 0.7459 & 70,542 & - \\
    \hline
    + FrameNet  & 0.7594 & 0.6323 & 0.4051 & 0.7740 & 7,637 & - \\
    \hlineB{3}
    \end{tabular}
    \caption{Spearman's correlation of extrofitted word vectors for word similarity tasks on pretrained word vectors by Word2Vec and Fasttext. Extrofitting can be applied to other kinds of pretrained word vector.}
    \label{tab:4}
    \end{table*}

\section{Experiments on Word Similarity Task}
    % We calculate cosine similarity between the vectors of two words in test item and report Spearman's rank correlation coefficient()
    The word similarity task is to calculate Spearman's correlation~\cite{daniel1990spearman} between two words as word vector format. We first apply extrofitting to GloVe from different data sources and present the result in Table~\ref{tab:1}.
    The result shows that although the number of the extrofitted word with FrameNet is less than the other lexicons, its performance is on par with other lexicons. We can also ensure that our method improves the performance of original pretrained word vectors.\\
    Next, we perform extrofitting on GloVe in different word dimension and compare the performance with retrofitting. We use WordNet$_{all}$ lexicon on both retrofitting and extrofitting to compare the performances in the ideal environment for retrofitting. We present the results in Table~\ref{tab:2}. We can demonstrate that our method outperforms retrofitting on some of word similarity tasks, MEN-3k and WordSim-353. We believe that extrofitting on SimLex-999 and RG-65 is less powerful because all word pairs in the datasets are included on WordNet$_{all}$ lexicon. Since retrofitting forces the word similarity to be improved by weighted averaging their word vectors, it is prone to be overfitted on semantic lexicons. On the other hand, extrofitting also uses synonyms to improve word similarity but it works differently that extrofitting projects the synonyms both close together on a new vector space and far from the other words. Therefore, our method can make more generalized word representation than retrofitting. We plot top-100 nearest words using t-SNE~\cite{maaten2008visualizing}, as shown in Figure~\ref{fig:1}. We can find that retrofitting strongly collects synonym words together whereas extrofitting weakly disperses the words, resulting loss in cosine similarity score. However, the result of extrofitting can be interpreted as generalization that the word vectors strengthen its own meaning by being far away from each other, still keeping synonyms relatively close together (see Table~\ref{tab:3}). When we list up top-10 nearest words, extrofitting shows more favorable results than retrofitting. We can also observe that extrofitting even can be applied to words which are not included in semantic lexicons.\\
    Lastly, we apply extrofitting to other well-known pretrained word vectors trained by different algorithms (see Subsection~\ref{pretrainedwordvectors}). The result is presented in Table~\ref{tab:4}. Extrofitting can be also applied to Word2Vec and Fasttext, enriching their word representations except on WordSim-353 and RG-65, respectively. We find that our method can distort the well-established word embeddings. However, our results are noteworthy in that extrofitting can be applied to other kinds of pretrained word vectors for further enrichment.
    
\section{Conclusion}
    We propose post-processing method for enriching not only word representation but also its vector space using semantic lexicons, which we call {\em extrofitting}. Our method takes a simple approach that (i) expanding word dimension (ii) transferring semantic knowledge on the word vectors (iii) projecting the vector space with enrichment. We show that our method outperforms another post-processing method, retrofitting, on some of word similarity task. Our method is robust in respect to the dimension of word vector and the size of vocabulary, only including an explainable hyperparameter; the number of dimension to be expanded. Further, our method does not depend on the order of synonym pairs. As a future work, we will do further research about our method to generalize and improve its performance; First, we can experiment on other word similarity datasets for generalization. Second, we can also utilize Autoencoder~\cite{bengio2009learning} for non-linear projection with a constraint of preserving spatial information of each dimension of word vector.
%Collapse multiple citations as
%in~\cite{Gusfield:97,Aho:72}. 

\section*{Acknowledgments}

Thanks for Jaeyoung Kim to discuss this idea. Also, greatly appreciate the reviewers for critical comments.
% The acknowledgments should go immediately before the references.  Do not number the acknowledgments section ({\em i.e.}, use \verb|\section*| instead of \verb|\section|). Do not include this section when submitting your paper for review.

%\bibliography{acl2018}
\bibliography{acl2018}

\begin{thebibliography}{23}
\expandafter\ifx\csname natexlab\endcsname\relax\def\natexlab#1{#1}\fi

\bibitem[{Baker et~al.(1998)Baker, Fillmore, and Lowe}]{baker1998berkeley}
Collin~F Baker, Charles~J Fillmore, and John~B Lowe. 1998.
\newblock The berkeley framenet project.
\newblock In \emph{Proceedings of the 17th international conference on
  Computational linguistics-Volume 1}, pages 86--90. Association for
  Computational Linguistics.

\bibitem[{Bengio et~al.(2009)}]{bengio2009learning}
Yoshua Bengio et~al. 2009.
\newblock Learning deep architectures for ai.
\newblock \emph{Foundations and trends{\textregistered} in Machine Learning},
  2(1):1--127.

\bibitem[{Bojanowski et~al.(2016)Bojanowski, Grave, Joulin, and
  Mikolov}]{bojanowski2016enriching}
Piotr Bojanowski, Edouard Grave, Armand Joulin, and Tomas Mikolov. 2016.
\newblock Enriching word vectors with subword information.
\newblock \emph{arXiv preprint arXiv:1607.04606}.

\bibitem[{Bruni et~al.(2014)Bruni, Tram, Baroni et~al.}]{bruni2014multimodal}
Elia Bruni, N~Tram, Marco Baroni, et~al. 2014.
\newblock Multimodal distributional semantics.
\newblock \emph{The Journal of Artificial Intelligence Research}, 49:1--47.

\bibitem[{Camacho-Collados et~al.(2015)Camacho-Collados, Pilehvar, and
  Navigli}]{camacho2015nasari}
Jos{\'e} Camacho-Collados, Mohammad~Taher Pilehvar, and Roberto Navigli. 2015.
\newblock Nasari: a novel approach to a semantically-aware representation of
  items.
\newblock In \emph{Proceedings of the 2015 Conference of the North American
  Chapter of the Association for Computational Linguistics: Human Language
  Technologies}, pages 567--577.

\bibitem[{Daniel(1990)}]{daniel1990spearman}
Wayne~W Daniel. 1990.
\newblock Spearman rank correlation coefficient.
\newblock \emph{Applied nonparametric statistics}, pages 358--365.

\bibitem[{Faruqui et~al.(2014)Faruqui, Dodge, Jauhar, Dyer, Hovy, and
  Smith}]{faruqui2014retrofitting}
Manaal Faruqui, Jesse Dodge, Sujay~K Jauhar, Chris Dyer, Eduard Hovy, and
  Noah~A Smith. 2014.
\newblock Retrofitting word vectors to semantic lexicons.
\newblock \emph{arXiv preprint arXiv:1411.4166}.

\bibitem[{Finkelstein et~al.(2001)Finkelstein, Gabrilovich, Matias, Rivlin,
  Solan, Wolfman, and Ruppin}]{finkelstein2001placing}
Lev Finkelstein, Evgeniy Gabrilovich, Yossi Matias, Ehud Rivlin, Zach Solan,
  Gadi Wolfman, and Eytan Ruppin. 2001.
\newblock Placing search in context: The concept revisited.
\newblock In \emph{Proceedings of the 10th international conference on World
  Wide Web}, pages 406--414. ACM.

\bibitem[{Ganitkevitch et~al.(2013)Ganitkevitch, Van~Durme, and
  Callison-Burch}]{ganitkevitch2013ppdb}
Juri Ganitkevitch, Benjamin Van~Durme, and Chris Callison-Burch. 2013.
\newblock Ppdb: The paraphrase database.
\newblock In \emph{Proceedings of the 2013 Conference of the North American
  Chapter of the Association for Computational Linguistics: Human Language
  Technologies}, pages 758--764.

\bibitem[{Hill et~al.(2015)Hill, Reichart, and Korhonen}]{hill2015simlex}
Felix Hill, Roi Reichart, and Anna Korhonen. 2015.
\newblock Simlex-999: Evaluating semantic models with (genuine) similarity
  estimation.
\newblock \emph{Computational Linguistics}, 41(4):665--695.

\bibitem[{Kiela et~al.(2015)Kiela, Hill, and Clark}]{kiela2015specializing}
Douwe Kiela, Felix Hill, and Stephen Clark. 2015.
\newblock Specializing word embeddings for similarity or relatedness.
\newblock In \emph{Proceedings of the 2015 Conference on Empirical Methods in
  Natural Language Processing}, pages 2044--2048.

\bibitem[{Maaten and Hinton(2008)}]{maaten2008visualizing}
Laurens van~der Maaten and Geoffrey Hinton. 2008.
\newblock Visualizing data using t-sne.
\newblock \emph{Journal of machine learning research}, 9(Nov):2579--2605.

\bibitem[{Mikolov et~al.(2013{\natexlab{a}})Mikolov, Chen, Corrado, and
  Dean}]{mikolov2013efficient}
Tomas Mikolov, Kai Chen, Greg Corrado, and Jeffrey Dean. 2013{\natexlab{a}}.
\newblock Efficient estimation of word representations in vector space.
\newblock \emph{arXiv preprint arXiv:1301.3781}.

\bibitem[{Mikolov et~al.(2013{\natexlab{b}})Mikolov, Sutskever, Chen, Corrado,
  and Dean}]{mikolov2013distributed}
Tomas Mikolov, Ilya Sutskever, Kai Chen, Greg~S Corrado, and Jeff Dean.
  2013{\natexlab{b}}.
\newblock Distributed representations of words and phrases and their
  compositionality.
\newblock In \emph{Advances in neural information processing systems}, pages
  3111--3119.

\bibitem[{Miller(1995)}]{miller1995wordnet}
George~A Miller. 1995.
\newblock Wordnet: a lexical database for english.
\newblock \emph{Communications of the ACM}, 38(11):39--41.

\bibitem[{Mrk{\v{s}}i{\'c} et~al.(2016)Mrk{\v{s}}i{\'c}, S{\'e}aghdha, Thomson,
  Ga{\v{s}}i{\'c}, Rojas-Barahona, Su, Vandyke, Wen, and
  Young}]{mrkvsic2016counter}
Nikola Mrk{\v{s}}i{\'c}, Diarmuid~O S{\'e}aghdha, Blaise Thomson, Milica
  Ga{\v{s}}i{\'c}, Lina Rojas-Barahona, Pei-Hao Su, David Vandyke, Tsung-Hsien
  Wen, and Steve Young. 2016.
\newblock Counter-fitting word vectors to linguistic constraints.
\newblock \emph{arXiv preprint arXiv:1603.00892}.

\bibitem[{Pedregosa et~al.(2011)Pedregosa, Varoquaux, Gramfort, Michel,
  Thirion, Grisel, Blondel, Prettenhofer, Weiss, Dubourg, Vanderplas, Passos,
  Cournapeau, Brucher, Perrot, and Duchesnay}]{scikit-learn}
F.~Pedregosa, G.~Varoquaux, A.~Gramfort, V.~Michel, B.~Thirion, O.~Grisel,
  M.~Blondel, P.~Prettenhofer, R.~Weiss, V.~Dubourg, J.~Vanderplas, A.~Passos,
  D.~Cournapeau, M.~Brucher, M.~Perrot, and E.~Duchesnay. 2011.
\newblock Scikit-learn: Machine learning in {P}ython.
\newblock \emph{Journal of Machine Learning Research}, 12:2825--2830.

\bibitem[{Pennington et~al.(2014)Pennington, Socher, and
  Manning}]{pennington2014glove}
Jeffrey Pennington, Richard Socher, and Christopher Manning. 2014.
\newblock Glove: Global vectors for word representation.
\newblock In \emph{Proceedings of the 2014 conference on empirical methods in
  natural language processing (EMNLP)}, pages 1532--1543.

\bibitem[{Rubenstein and Goodenough(1965)}]{rubenstein1965contextual}
Herbert Rubenstein and John~B Goodenough. 1965.
\newblock Contextual correlates of synonymy.
\newblock \emph{Communications of the ACM}, 8(10):627--633.

\bibitem[{Speer et~al.(2017)Speer, Chin, and Havasi}]{speer2017conceptnet}
Robert Speer, Joshua Chin, and Catherine Havasi. 2017.
\newblock Conceptnet 5.5: An open multilingual graph of general knowledge.
\newblock In \emph{AAAI}, pages 4444--4451.

\bibitem[{Turian et~al.(2010)Turian, Ratinov, and Bengio}]{turian2010word}
Joseph Turian, Lev Ratinov, and Yoshua Bengio. 2010.
\newblock Word representations: a simple and general method for semi-supervised
  learning.
\newblock In \emph{Proceedings of the 48th annual meeting of the association
  for computational linguistics}, pages 384--394. Association for Computational
  Linguistics.

\bibitem[{Vuli{\'c} et~al.(2017)Vuli{\'c}, Mrk{\v{s}}i{\'c}, and
  Korhonen}]{vulic2017cross}
Ivan Vuli{\'c}, Nikola Mrk{\v{s}}i{\'c}, and Anna Korhonen. 2017.
\newblock Cross-lingual induction and transfer of verb classes based on word
  vector space specialisation.
\newblock \emph{arXiv preprint arXiv:1707.06945}.

\bibitem[{Welling(2005)}]{welling2005fisher}
Max Welling. 2005.
\newblock Fisher linear discriminant analysis.
\newblock \emph{Department of Computer Science, University of Toronto}, 3(1).

\end{thebibliography}
\bibliographystyle{acl_natbib}

\end{document}